\pdfoutput=1
\documentclass[runningheads]{llncs}

\RequirePackage{silence}
\usepackage[T1]{fontenc}
\usepackage[utf8]{inputenc}
\usepackage{graphicx}
\usepackage{amsmath,amssymb}
\usepackage{color}
\usepackage{xcolor}
\usepackage{url}
\usepackage{hyperref}
\usepackage{xspace}
\usepackage{booktabs}
\usepackage{multirow}
\usepackage{array}
\usepackage{caption}
\usepackage{subcaption}
\usepackage{enumitem}
\usepackage{placeins}

\definecolor{ForestGreen}{RGB}{34,139,34}
\definecolor{Dandelion}{RGB}{253,188,66}
\definecolor{mypurple}{RGB}{150,110,200}
\definecolor{myorange}{RGB}{245,165,60}
\definecolor{myblue}{RGB}{25,92,160}
\definecolor{mygreen}{RGB}{120,200,120}

\newcommand{\etal}{\emph{et al.}\@\xspace}

\newcommand{\titledescription}{Language-Guided Semantic Shape Editing of 3D Gaussian Head Avatars}
\newcommand{\method}{ManifoldSplat\xspace}
\newcommand{\shapemodel}{DeltaRegion\xspace}

\def\SubNumber{14}
\def\GCPRTrack{Main Track}

\newif\ifreview
\reviewfalse

\ifreview
    \usepackage{lineno}

    \linenumbers
\fi

\begin{document}

% -----------------------------------------------------------------------------
% Submission metadata
% -----------------------------------------------------------------------------

\title{\method: \titledescription}

\ifreview
    \titlerunning{GCPR 2026 Submission \SubNumber{}. CONFIDENTIAL REVIEW COPY.}
    \authorrunning{GCPR 2026 Submission \SubNumber{}. CONFIDENTIAL REVIEW COPY.}
    \author{GCPR 2026 - \GCPRTrack{}}
    \institute{Paper ID \SubNumber}
\else
    % CAMERA READY SUBMISSION

    \titlerunning{\method: Language-Guided Semantic Shape Editing}
    \authorrunning{A. Canela and J. S\`anchez-Riera}

    % \author{Antonio Canela\inst{1,2}\orcidID{0009-0002-1943-5170} \and
    % Jordi S\`anchez-Riera\inst{2}\orcidID{0000-0002-4803-5742}}

	\author{Antonio Canela\inst{1,2}\thanks{Corresponding Author} \and
    Jordi S\`anchez-Riera\inst{2}}

	\institute{Universitat Polit\`ecnica de Catalunya \and
    Institut de Rob\`otica i Inform\`atica Industrial, CSIC-UPC\\
    \email{antonio.canela@upc.edu, jsanchez@iri.upc.edu}}
\fi

\maketitle

% -----------------------------------------------------------------------------
% Preliminary sections
% -----------------------------------------------------------------------------

{
	% \vspace{-0.05cm}
	\centering
	\includegraphics[width=\textwidth]{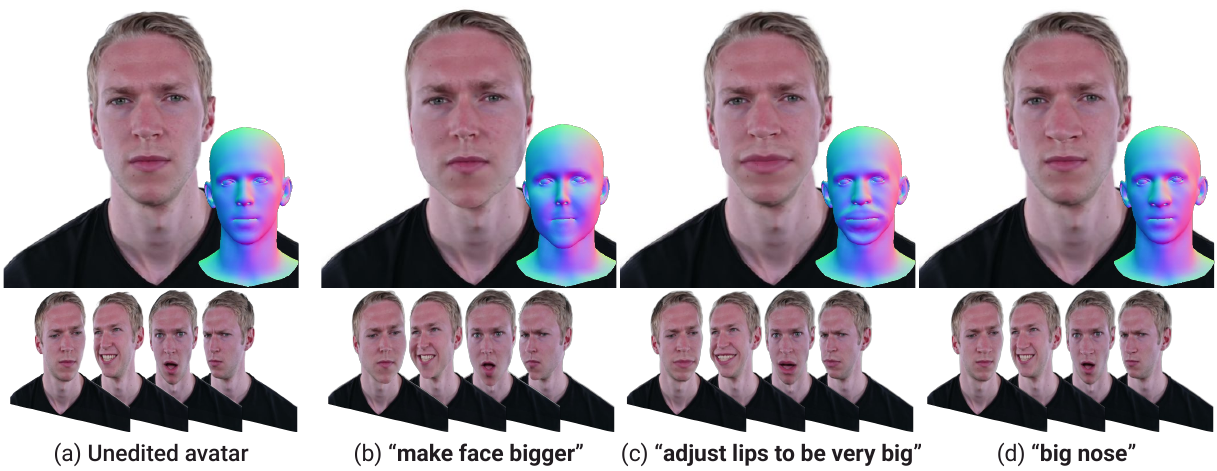}
	\vspace{-0.4cm}
	\captionsetup{type=figure}
	\caption{\textbf{Text-driven editing of animatable head avatars.} Our method produces localized, semantically meaningful shape changes through natural language prompts while preserving identity, photorealism, and animatability.}
	%  \textcolor{orange}{puedes cambiar un poco el texto del teaser? 1. un poco mas grande, y el texto un poco diferente. Por ejemplo, face much bigger me imagino una cabeza super grande, pero visualmente parece solo un poco mas grande, igual puedes poner "make face bigger", lo mismo para el significantly big nose, el edit no me parece tan granda cuando por el texto me imagino algo super grande}
	% \caption{\textbf{Text-driven editing of animatable head avatars.} From a single unedited avatar (a), \method applies large, semantically distinct shape edits (b-d) while preserving identity, photorealism and the expression rig (bottom rows).}
	\label{fig:teaser}
	% \vspace{-0.1cm}
	\par
}

\begin{abstract}

\vspace{-0.2cm}

High-fidelity 3D head avatars have reached near-photorealistic quality. While recent methods enable text-driven manipulation, they struggle to provide fine-grained localized control, often entangling features or lacking geometric consistency. Modifying geometry through natural language currently requires slow per-prompt optimization or compromises identity and rigging. We present \method, the first end-to-end framework for language-guided semantic shape editing of animatable 3D Gaussian Splatting avatars reconstructed from monocular videos. By performing edits within the structured FLAME manifold rather than directly optimizing an unstructured Gaussian cloud, we strictly preserve identity and animation. We introduce \shapemodel, a per-region disentangled Conditional Variational Autoencoder (CVAE) delivering feed-forward shape deltas, alongside a refining stage to recover view-consistent details. \method reconstructs and edits an avatar in $\sim$\,90~seconds on a consumer GPU, rendering at $\sim$\,800\,FPS. Extensive evaluations demonstrate our approach sets a new state-of-the-art in localized prompt alignment, geometric coherence, and identity preservation. Project page and code: \url{https://a-canela.github.io/manifoldsplat/}

\vspace{-0.2cm}

\end{abstract}


% -----------------------------------------------------------------------------
% Main sections
% -----------------------------------------------------------------------------

\section{Introduction}
\label{sec:intro}

Photorealistic animatable 3D head avatars are a core building block of
telepresence, virtual production, and assistive technologies. By
combining a parametric face model such as \textsc{FLAME}~\cite{flame17}
with 3D Gaussian Splatting~\cite{kerbl3Dgaussians}, recent
methods~\cite{qian2024gaussianavatars,xiang2024flashavatar,shao2024splattingavatar,HRAvatar,li2025rgbavatar}
can reconstruct a personal, drivable head from a short monocular video
in minutes and render it in real time. Once reconstructed, however,
these avatars are frozen identities: there is no built-in way
for a user to manipulate specific facial attributes through text
and obtain, in seconds, an animatable avatar of the same person with
the requested change (Fig.~\ref{fig:teaser}).

Closing this gap remains challenging. Text-to-3D-face
generators~\cite{wu2023describe3d,aneja2023clipface,semantify} synthesize
meshes from scratch, failing to preserve personal identity.
Distillation-based editors~\cite{brooks2022instructpix2pix,liu2025gaussianavatar-editor}
require lengthy per-prompt optimization, focus strictly on appearance,
and often break multi-view consistency. Meanwhile, generic 3D scene
editors~\cite{gaussianeditor,wang2024gaussianeditor2} demand multi-view captures.
Crucially, no existing method achieves localized geometric editing
of a personal monocular avatar while preserving its animation rigging
and sub-minute latency.

Our key observation is that \textsc{FLAME}'s identity subspace is
already a structured manifold of plausible head geometry, but it lacks
explicit correspondence with individual facial attributes, making
isolated edits inherently difficult. Learning a text-conditioned
generator inside this manifold - rather than in the Gaussian or
NeRF parameter space - yields edits that are by construction
physically plausible, identity-preserving, and trivially compatible
with downstream animation. The remaining challenges are then to
localize an edit to the region the user names without entangling
unrelated attributes, and to make the Gaussian avatar robust to the
resulting geometric change without a full per-edit optimization.

We propose \method, a two-stage pipeline on a mesh-bound 3D Gaussian
Splatting avatar: a text-guided shape-editing stage produces a
localized identity delta on the \textsc{FLAME} manifold, followed by a
targeted rendering stage that adapts the splats so the
re-parameterized identity renders cleanly (Sec.~\ref{sec:method}).

In our evaluations, \method outperforms
GaussianAvatar-Editor~\cite{liu2025gaussianavatar-editor} and
InstructPix2Pix~\cite{brooks2022instructpix2pix} on prompt alignment,
identity preservation and visual cleanliness while being one to two
orders of magnitude faster; at the shape level, it significantly
outperforms \mbox{DESCRIBE3D}~\cite{wu2023describe3d} and
Semantify~\cite{semantify} in edit locality.

\paragraph{Contributions.}
\begin{itemize}
    \item \textbf{\method:} an end-to-end pipeline
    achieving state-of-the-art language-guided semantic shape editing
    of animatable 3D Gaussian head avatars reconstructed from
    monocular video, with targeted rendering
    enhancements that address the characteristic geometric artifacts.

    \item \textbf{\shapemodel:} a text-guided shape editing model
    that, given a free-form prompt and a source \textsc{FLAME}
    identity, yields a localized shape delta in a single feed-forward
    pass.

    \item \textbf{A synthetic dataset} of per-region
    vertex-displacement edits defined on the \textsc{FLAME}
    topology.
\end{itemize}

\section{Related Work}
\label{sec:relatedwork}

\method sits at the intersection of multiple lines of research: parametric
3D face models, animatable head avatars with neural and Gaussian
representations, and text-driven 3D generation and editing.

\subsection{Parametric Face Models.}
Linear morphable face models~\cite{lyhm,flame17} and the encoder-based
identity predictors built on them~\cite{mica_eccv22} derive their shape
spaces via global dimensionality reduction, so their parameters are
highly entangled and lack explicit correspondence with localized
semantic attributes. While mapping linguistic descriptions to global
shape spaces has been explored for parametric body
models~\cite{streuber2016bodytalk}, achieving localized, disentangled
control over specific facial attributes on the \textsc{FLAME} manifold
remains an open challenge. \shapemodel addresses this gap by operating
entirely inside the \textsc{FLAME} identity shape space, giving us a
strong inductive bias for plausibility while enforcing localized,
semantic control, and keeping every edit trivially compatible with
downstream \textsc{FLAME}-rigged assets.

\subsection{Head-Avatar Reconstruction.}
Early monocular head avatars relied on neural radiance
fields~\cite{mildenhall2020nerf,nerface,zielonka2022insta,Zheng2023pointavatar}.
Driven by its real-time efficiency and high visual fidelity, 3D
Gaussian Splatting~\cite{kerbl3Dgaussians} shifted the dominant
paradigm toward explicit, animatable Gaussian representations rigged
on a parametric
mesh~\cite{qian2024gaussianavatars,shao2024splattingavatar,xiang2024flashavatar,HRAvatar,li2025rgbavatar};
a parallel branch tackles cross-identity, feed-forward avatar
synthesis from one or a few images~\cite{chu2024gagavatar,he2024head360,avat3r}.
We build on the Gaussian-blendshape formulation
of~\cite{li2025rgbavatar} because its identity-shape-driven
parameterization makes shape edits both fast and well defined; our
rendering-side enhancements (Sec.~\ref{sec:method:enhancements}) are
nonetheless orthogonal and could be integrated into other
mesh-anchored Gaussian avatars.

\subsection{Text-Driven 3D Face Generation.}
A growing body of work generates 3D faces or full heads from a
text prompt, either via Score Distillation
Sampling~\cite{poole2023dreamfusion} on top of generic text-to-3D
backbones~\cite{lin2023magic3d,tang2024dreamgaussian} or with
face-specific pipelines~\cite{wu2023describe3d,dreamface,aneja2023clipface,zhou2024headstudio,liao2024tada,liu2024headartist};
CLIP-space stylization
methods~\cite{michel2022text2mesh,wang2022clipnerf} explore a related
formulation. All of these target generation rather than
editing of an identity reconstructed from a user-supplied
video, so comparing \method against them would conflate two problem
statements. To still isolate the shape-editing aspect of our pipeline,
we benchmark \shapemodel against \mbox{DESCRIBE3D}~\cite{wu2023describe3d}
and Semantify~\cite{semantify} as the closest practical shape-level
counterparts, noting that both synthesize from a neutral mesh without
conditioning on a source identity.

\subsection{Editing 3D Avatars and Scenes.}
Text-driven scene editing was popularized by
InstructPix2Pix~\cite{brooks2022instructpix2pix} and its 3D lifts
Instruct-NeRF2NeRF and
Instruct-GS2GS~\cite{haque2023instructnerf2nerf,vachha2024instruct-gs2gs},
followed by multi-view-consistent
editors~\cite{gaussianeditor,wang2024gaussianeditor2,chen2024dge}.
Head-avatar-specific methods include
GaussianAvatar-Editor~\cite{liu2025gaussianavatar-editor} (per-prompt
InstructPix2Pix iteration on an animatable 3DGS head),
FATE~\cite{zhang2025fate} (textural edits),
AvatarStudio~\cite{mendiratta2023avatarstudio} (dynamic-actor editing
via 2D diffusion priors), and ShapeFusion~\cite{potamias2024shapefusion}
(localized shape edits via a 3D diffusion model conditioned on sparse
handles). Most multi-view scene editors require COLMAP-style
calibration or assume a static scene, both violated by monocular
talking-head video. The only published methods that meet our
constraints (animatable, monocularly reconstructed Gaussian head,
free-form text instruction) are
GaussianAvatar-Editor~\cite{liu2025gaussianavatar-editor} and an
InstructPix2Pix-based editor of the trained avatar, which we use as
our two avatar-level baselines.

% =============================================================================
\begin{figure}[htbp]
  \vspace{-0.5cm}
  \centering
  \includegraphics[width=\textwidth]{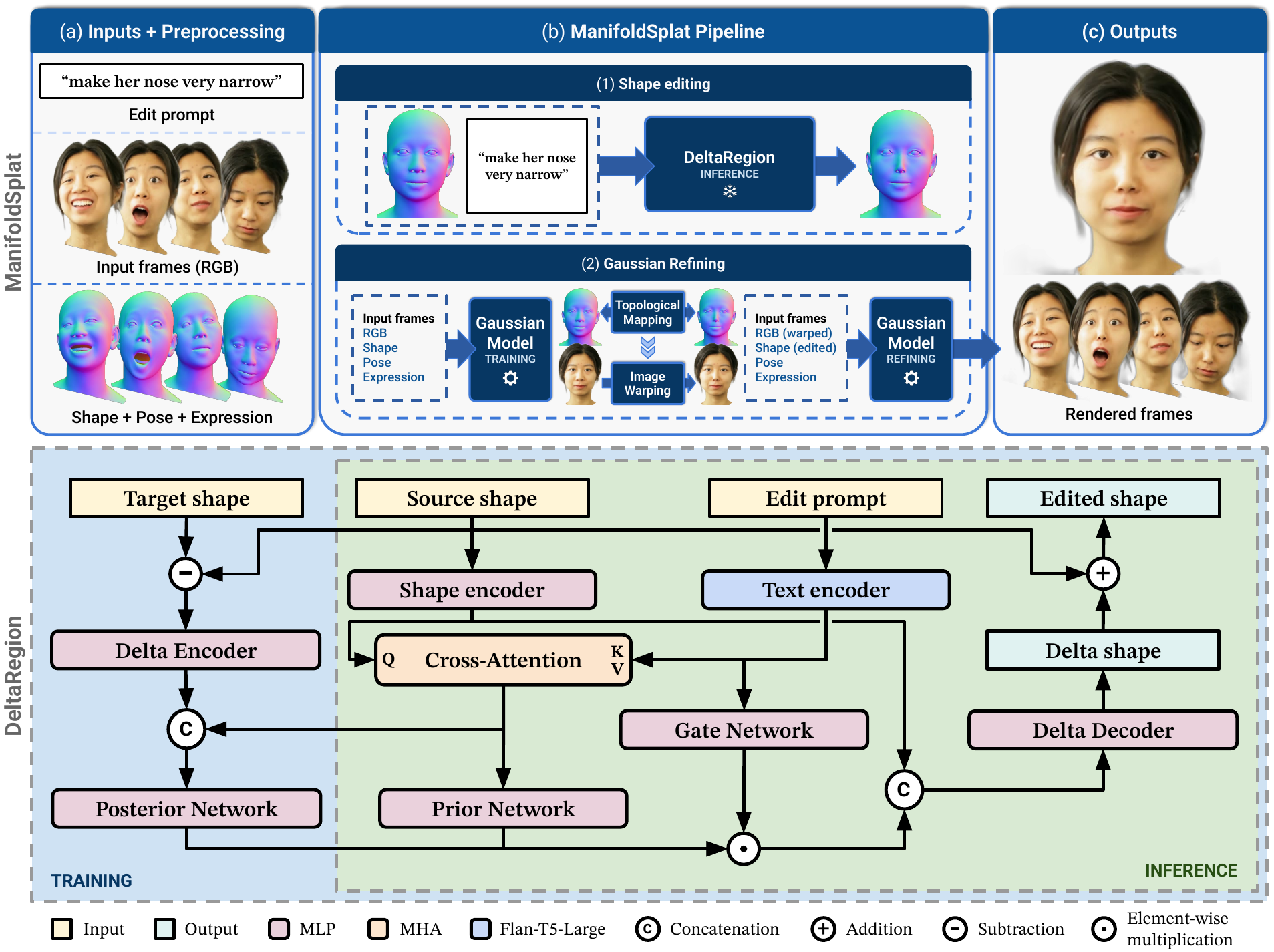}
  \caption{\textbf{Overview of \method.} \emph{Top:} the \method
  pipeline turns a monocular video and a prompt into an animatable edited
  avatar via a frozen \shapemodel edit followed by Gaussian training and
  warped refining (Sec.~\ref{sec:method:enhancements}). \emph{Bottom:}
  \shapemodel, our text-conditioned shape-editing CVAE; the region gate
  zeroes out non-selected latent blocks before decoding
  (Sec.~\ref{sec:method:shapemodel}).}
  \label{fig:method}
  \vspace{-1.0cm}
\end{figure}
% =============================================================================

\section{Method}
\label{sec:method}

\method takes a short monocular RGB video $\mathcal{V}$ of a subject and a free-form prompt $\mathbf{t}$ (e.g.\ ``shorter nose''), and produces an animatable 3D Gaussian Splatting head avatar with a semantically edited \textsc{FLAME} identity. As overviewed in Fig.~\ref{fig:method}, our pipeline bridges parametric mesh edits and splat-based rendering in two stages: \emph{(i)} text-driven shape editing via \shapemodel for a localized geometric edit on the \textsc{FLAME} manifold (Sec.~\ref{sec:method:shapemodel}), and \emph{(ii)} a targeted Gaussian refining stage that adapts splat parameters to the new geometry (Sec.~\ref{sec:method:enhancements}).

\subsection{Background - Preprocessing and Mesh-bound Backbone}
\label{sec:method:background}

\noindent\textbf{Preprocessing.} We adopt
RGBAvatar~\cite{li2025rgbavatar} as our reconstruction backbone and
inherit its preprocessing stack: MICA~\cite{mica_eccv22} for
identity initialization, metrical-tracker~\cite{mica_eccv22} for
per-frame \textsc{FLAME} pose and expression, and Robust Video
Matting~\cite{RobustVideoMatting} for foreground extraction at
$512{\times}512$.

\noindent\textbf{FLAME}~\cite{flame17} parameterizes a head mesh as
$M(\boldsymbol{\beta},\boldsymbol{\psi},\boldsymbol{\theta})$ with
identity $\boldsymbol{\beta}\!\in\!\mathbb{R}^{300}$, expression
$\boldsymbol{\psi}$, and pose $\boldsymbol{\theta}$ in separate linear
subspaces, so editing $\boldsymbol{\beta}$ alone changes \emph{who} the
subject is without altering \emph{how} they move.

\noindent\textbf{Mesh-bound Gaussian-blendshape avatars} represent a
personal head as a set of 3D Gaussians whose canonical centers are
anchored on the \textsc{FLAME} mesh and deformed by a small set of $B$
learned linear blendshapes driven by the tracker's expression vector.
Each Gaussian $g_i$ stores a rest position $\boldsymbol{x}^{(0)}_i$
expressed in the tangent-bitangent-normal (TBN) frame of its
anchor triangle, a set of per-blendshape offsets
$\{\boldsymbol{x}^{(b)}_i\}_{b=1}^{B}$ in the same local frame, plus
rotation, scale, opacity and spherical-harmonic coefficients (DC band
in our setting). Triangle-local parameterization is key: any deformation of the underlying mesh - including a change
of identity $\boldsymbol{\beta}$ - is automatically propagated to the
Gaussians through the triangle TBN, bypassing full re-optimization.

% =============================================================================
\subsection{Synthetic Dataset}
\label{sec:method:dataset}

Training \shapemodel requires a large pool of triplets $(\boldsymbol{\beta}_0,
\mathbf{t}, \boldsymbol{\beta}_1)$, each pairing a source identity
$\boldsymbol{\beta}_0$ and an edit prompt $\mathbf{t}$ with the target
identity
$\boldsymbol{\beta}_1=\boldsymbol{\beta}_0+\Delta\boldsymbol{\beta}$
realizing the prompted edit, densely covering both the
identity manifold and the space of region-localized semantic edits. No
public corpus pairs free-form prompts with identity-preserving
\textsc{FLAME} shape deltas, and hand-annotating target meshes per
subject does not scale. Our key design choice is to specify each edit
once as a sparse set of 3D vertex displacements on the canonical
\textsc{FLAME} mesh, rather than as a precomputed shape-coefficient
target. Since all \textsc{FLAME} identities share the same topology,
the same prototype can be projected onto any source identity
$\boldsymbol{\beta}_0$ by solving for the $\Delta\boldsymbol{\beta}$
whose induced vertex motion best matches the prescribed displacements
on that subject's deformed mesh, turning a small library of
prototypes into a combinatorially large supervised
set without per-subject mesh annotation.

\paragraph{Scalable Generation.}
We author $\sim$130 fine-grained prototypes spanning the six face
regions (\emph{face, nose, lips, ears, neck, eyes}); each is a JSON
descriptor that picks a small set of \textsc{FLAME} vertex indices in
the targeted region, assigns each a 3D displacement vector, and
provides a canonical prompt fragment (e.g.\ \emph{``wide nose
bridge''}). Source identities come from MICA-fitted
LYHM~\cite{lyhm} subjects; for every (subject, edit, scale) tuple we
recover $\Delta\boldsymbol{\beta}$ by minimizing the squared distance
between the deformed mesh and the displaced anchor vertices under an
outside-region penalty that prevents leakage into unrelated areas,
with displacements linearly rescaled per tuple to vary edit magnitude.
Each tuple is then paired with several prompt variations, generated by
injecting scale-dependent qualifiers into templates built around the
canonical fragment and paraphrasing via Flan-T5~\cite{chung2024flan,raffel2020t5} for linguistic
diversity. This single pass yields $\sim$500k supervised triplets
(details in supplementary).

% =============================================================================
\subsection{Stage 1 - Shape Editing}
\label{sec:method:shapemodel}

\paragraph{Problem.}
Given a prompt $\mathbf{t}$ and a source identity
$\boldsymbol{\beta}_0$, we want
$\boldsymbol{\beta}_1=\boldsymbol{\beta}_0+\Delta\boldsymbol{\beta}$ such
that (i) the edit matches $\mathbf{t}$ semantically, (ii) it is
local - it should not spill into facial regions the prompt does
not mention - and (iii) the model exposes a continuous knob for edit
strength while remaining deterministic enough to be reproducible.
We cast this as a conditional generative model mapping the
(prompt, source-shape) pair to the edited
\textsc{FLAME} shape.

\paragraph{Architecture.}
\shapemodel (Fig.~\ref{fig:method}, bottom) is a conditional VAE that
fuses a frozen Flan-T5~\cite{chung2024flan,raffel2020t5} text encoding
with a small MLP embedding of $\boldsymbol{\beta}_0$ via
cross-attention, then samples a structured latent
$\mathbf{z}\!\in\!\mathbb{R}^{Kd}$ split into $K{=}6$ contiguous
blocks of dimension $d$, one per face region. A small gate network predicts a softmax distribution
$\mathbf{g}\!\in\!\Delta^{K-1}$ over regions from the prompt
embedding alone, and each block is multiplied by its gate
value before being decoded alongside the source shape embedding into the shape delta
$\Delta\boldsymbol{\beta}\!\in\!\mathbb{R}^{300}$. This gate explicitly enforces disentangled
editing: any latent dimension belonging to a non-selected region is
multiplied by a near-zero value before reaching the decoder, so its
contribution to $\Delta\boldsymbol{\beta}$ vanishes
by construction. Full layer sizes, the cross-attention block
and the prior/posterior parameterization are listed in the
supplementary material.

\paragraph{Training.}
We train \shapemodel on the corpus of
Sec.~\ref{sec:method:dataset} under the combined objective
\begin{equation}
  \mathcal{L} \;=\;
    \lambda_{v}^{\text{in}}\,\mathcal{L}_{v}^{\text{in}} +
    \lambda_{v}^{\text{out}}\,\mathcal{L}_{v}^{\text{out}} +
    \lambda_{\beta}\,\mathcal{L}_{\beta} +
    \lambda_{g}\,\mathcal{L}_{\text{gate}} +
    \lambda_{\text{KL}}\,\mathcal{L}_{\text{KL}}^{\text{fb}},
  \label{eq:shapemodel_loss}
\end{equation}
Here $a^{*}\!\in\!\{1,\dots,K\}$ denotes the ground-truth region label of each training triplet and $V_{a^{*}}$ the set of \textsc{FLAME} vertices belonging to that region. The loss combines two region-masked vertex reconstruction terms: $\mathcal{L}_{v}^{\text{in}}$, which optimizes the vertices inside the ground-truth region mask $V_{a^{*}}$ toward the target shape, and $\mathcal{L}_{v}^{\text{out}}$, which anchors the vertices outside the mask to the unedited source shape. Additionally, the objective includes a shape-parameter $\ell_2$ penalty $\mathcal{L}_{\beta}$, a cross-entropy loss $\mathcal{L}_{\text{gate}}$ for the predicted region probabilities against $a^{*}$, and a free-bits KL divergence regularization $\mathcal{L}_{\text{KL}}^{\text{fb}}$~\cite{kingma2014vae} with linear warm-up. The $\lambda$ factors are scalar weights balancing the terms; we use $\lambda_{v}^{\text{in}}{=}\lambda_{v}^{\text{out}}{=}5\!\times\!10^{5}$ (compensating the metric scale of vertex coordinates), $\lambda_{\beta}{=}10$, $\lambda_{g}{=}10$ and $\lambda_{\text{KL}}{=}0.05$. The joint supervision of the masked vertex and gate terms is what yields the per-region disentanglement observed at inference (Fig.~\ref{fig:shape_vs_baselines}); full term definitions are detailed in the supplementary material.

\paragraph{Inference.}
At test time we drop the posterior and decode the mean
$\boldsymbol{\mu}_p$ of the learned conditional prior
$p_\theta(\mathbf{z}\!\mid\!\mathbf{t},\boldsymbol{\beta}_0)=
\mathcal{N}(\boldsymbol{\mu}_p,\boldsymbol{\sigma}_p^2)$, rescaled
by a \emph{prior-scale factor} $s$,
$\mathbf{z}=s\,\boldsymbol{\mu}_p$, which compensates for the natural
attenuation of $\boldsymbol{\mu}_p$ relative to training-time
posterior samples; we set $s{=}1.2$ empirically on a held-out
validation split. The decoder
then yields $\Delta\boldsymbol{\beta}$ in one feed-forward pass, and
the mapping is deterministic per prompt. Edit strength is exposed
through two complementary continuous knobs: linguistically, the
scale-dependent qualifiers in the prompt (\emph{``slightly''},
\ldots, \emph{``drastically''}), which the model learns to associate
with displacement magnitude from the scale-varied training tuples;
and numerically, the prior-scale factor $s$ itself, which can be
varied continuously at test time to smoothly attenuate or exaggerate
the decoded delta.

% =============================================================================
\subsection{Stage 2 - Gaussian Refining}
\label{sec:method:enhancements}

Once the shape has been edited by \shapemodel, plugging the new identity
$\boldsymbol{\beta}_1$ into an off-the-shelf mesh-bound avatar reveals
multiple failure modes: Gaussians that were comfortable on the original surface
now hover above or sink below the edited mesh (floaters), high-frequency color
transitions become jagged because the local Gaussian neighborhood has shifted,
and persistent appearance cues that the avatar had quietly memorised on
the original geometry - a mole, a stray eyebrow hair, the rim of an
iris - now sit in the wrong place. We address each with a targeted
change to the training pipeline. Naive local geometric or colorimetric
smoothing could not play this role: these artifacts are
systematic misalignments between splats and the edited surface
rather than zero-mean noise, so local averaging would leave the
misplaced cues in place while irreversibly blurring the
high-frequency detail (skin pores, moles, iris rims) that
photorealism depends on; our objectives instead re-align that
detail with the new geometry. These changes do not alter the avatar's
representation or renderer interface; they only affect how Gaussians
are parameterized, activated, and re-optimized.

Temporally, this stage consists of two sequential optimizations.
First, the base avatar is reconstructed once per subject with
the surface-consistency regularizer (a) and the smooth color
activation (b) enabled: $\mathcal{L}_{\text{surf}}$ is added to the
backbone's photometric objective, and all terms are minimized
jointly. Then, after \shapemodel delivers $\boldsymbol{\beta}_1$,
the short warped-refining pass (c) restarts from this base checkpoint
(shared across all edits of the same subject) and simultaneously
minimizes a single combined objective - comprising the
photometric loss on the warped pseudo-ground-truths,
$\mathcal{L}_{\text{surf}}$, and the anchor term defined in Equation~\ref{eq:anchor}.

\paragraph{(a) Surface-consistency regularization.}
Because each Gaussian's rest position $\boldsymbol{x}^{(0)}_i$ and its
per-blendshape offsets $\{\boldsymbol{x}^{(b)}_i\}$ are expressed in
the TBN frame of the anchor triangle, an L2 penalty on these vectors
is exactly a penalty on deviation from the mesh surface:
\begin{equation}
  \mathcal{L}_{\text{surf}} \;=\;
    \lambda_{x}\sum_i\|\boldsymbol{x}^{(0)}_i\|_2^2 \;+\;
    \lambda_{\Delta x}\sum_{i,b}\|\boldsymbol{x}^{(b)}_i\|_2^2.
  \label{eq:surf}
\end{equation}
Penalizing both terms, with $\lambda$ factors as weights, pulls splats
onto the canonical surface and keeps
their per-blendshape motion coherent with the mesh, ensuring Gaussians
follow when the mesh is later re-parameterized by a shape edit.

\paragraph{(b) Smooth color activation.}
The backbone maps spherical-harmonic coefficients $\mathbf{h}$ to
colors through a sigmoid-like activation that saturates abruptly and
produces sharp transitions where neighboring Gaussians disagree - an
effect exacerbated when an edit perturbs the local Gaussian
neighborhood. We replace it inside the CUDA rasterization kernel by
$c = \tfrac12(\tanh(v)+1)$, where $v$ is the unbounded color value
obtained by evaluating $\mathbf{h}$ from the current view direction. The
mapping has the same $[0,1]$ range but a milder slope around zero and
a symmetric saturation profile, yielding smoother color fields
under shape edits at zero rendering cost.

\paragraph{(c) Warped refining.}
Even with (a) and (b), simply swapping $\boldsymbol{\beta}_0\!\to\!
\boldsymbol{\beta}_1$ cannot recover the view-dependent detail that the
trained avatar had implicitly encoded against the unedited geometry.
We therefore add a short warped-refining stage that exploits the
shared \textsc{FLAME} topology to compute, in closed form, a per-pixel
resampling map $W_t$ from each training-frame render $\tilde I_t$ of the unedited
mesh to the rasterization of the edited mesh: the source pixel
is found via triangle-and-barycentric reprojection. The view-consistent
pseudo-ground-truth $\hat I_t = W_t(\tilde I_t)$ follows from a single
texture-space reprojection. Because the warp is derived from the shared mesh
topology rather than from image-space optical flow, it is insensitive
to texture and free of the bracketing failures of optical-flow
estimators on low-texture regions like cheeks or forehead.

We then re-optimize the avatar on $\{\hat I_t\}$ with the edited
\textsc{FLAME} shape, the color channels frozen (so the warped
appearance serves as a strict chromatic reference), the
surface-consistency regularizer of Eq.~\ref{eq:surf} still active, and
an anchor term that prevents the non-frozen parameters $\Phi$
(positions, opacities, scales, rotations and the position/rotation
blendshape bases) from drifting away from the warm start
$\Phi^{(0)}$:
\begin{equation}
  \mathcal{L}_{\text{anchor}} \;=\;
    \sum_{\phi\in\Phi}\|\phi-\phi^{(0)}\|_2^2.
  \label{eq:anchor}
\end{equation}
Freezing color channels is what turns refining into a
geometric correction: positions, scales and rotations realign
with the new surface while the warped image dictates where pre-existing
appearance cues should sit, which is what removes the ghosting
artifacts of bare re-parameterization. Coupled with the anchor loss,
this stage produces a tightly local edit that preserves the unedited
avatar's overall identity and rendering quality everywhere outside the
region the prompt targets.

\section{Experiments}
\label{sec:experiments}

In the absence of existing baselines that perform localized geometric editing on monocular animatable avatars, we decouple our evaluation into three complementary tasks:
\emph{(i)} the full avatar editing task, comparing the complete
\method pipeline against
GaussianAvatar-Editor~\cite{liu2025gaussianavatar-editor} and an
InstructPix2Pix-based~\cite{brooks2022instructpix2pix} editor of the
trained avatar; \emph{(ii)} the shape-only generation task in which \shapemodel is
compared qualitatively against \mbox{DESCRIBE3D}~\cite{wu2023describe3d} and
Semantify~\cite{semantify} on a neutral \textsc{FLAME} identity;
and \emph{(iii)} an ablation study isolating the
\textsc{3DGS} enhancements (Surface Consistency, Smooth-Color tanh,
Warped Refining).

% =============================================================================

\subsection{Experimental Setup}
\label{sec:experiments:setup}

For avatar editing evaluation, we use the publicly-available, pre-processed subset of
\textsc{NeRSemble}~\cite{nersemble} released by
GaussianAvatar-Editor~\cite{liu2025gaussianavatar-editor}. Each subject's
front camera is held out for evaluation, with $100$-$300$ test frames per
(subject, edit) pair. We evaluate $18$ natural-language edits per subject
covering all six attribute regions. The evaluation prompts are not
verbatim copies of the training tuples: they introduce small
linguistic variations in wording and qualifiers over the canonical
fragments, which the frozen Flan-T5 encoder maps robustly to the
intended edit. We note, however, that robustness to strongly
out-of-distribution phrasings is not the focus of this work, whose
emphasis lies on the geometric locality and avatar consistency of
the edits.
All experiments were run on a single NVIDIA RTX~5090. \shapemodel is trained for $100$ epochs, taking $\sim$\,$40$\,h. At test time, per-subject avatar reconstruction runs for one minute with our enhancements enabled, while the optional warped refining stage requires an additional $30$ seconds.
To evaluate our method, we report three complementary metrics: CLIP-D~\cite{clip,gal2021stylegannada} for prompt alignment, ArcFace~\cite{deng2019arcface} for identity preservation, and Anisotropic 2D Total Variation ($TV_{2D}$) as a no-reference image-quality proxy. Detailed metric definitions are provided in the supplementary material.

% =============================================================================

\subsection{Quantitative results}
\label{sec:experiments:quantitative}

\begin{table}[htbp]
  \vspace{-0.7cm}
  \centering
  \caption{\textbf{Full-avatar editing quantitative comparison.}}
  \label{tab:avatar}
  \begin{tabular*}{\columnwidth}{@{\extracolsep{\fill}}lccc@{}}
    \toprule
    Method & CLIP-D $\uparrow$ & ArcFace $\uparrow$ & $TV_{2D}$ $\downarrow$ \\
    \midrule
    InstructPix2Pix~\cite{brooks2022instructpix2pix} & 0.0432 & 0.5212 & 0.0210 \\
    GaussianAvatar-Editor~\cite{liu2025gaussianavatar-editor} & 0.0467 & 0.7600 & 0.0126 \\
    \textbf{\method (Ours)} & \textbf{0.0490} & \textbf{0.9319} & \textbf{0.0122} \\
    \bottomrule
  \end{tabular*}
  \vspace{-0.8cm}
\end{table}

\paragraph{Full-avatar editing.}
To evaluate the full pipeline, we report the average CLIP-D, ArcFace, and $TV_{2D}$ (Table~\ref{tab:avatar}). \method tops every metric. The CLIP-D gap
($+5\%$ over GaussianAvatar-Editor, $+13\%$ over InstructPix2Pix) is
modest because CLIP-D rewards any plausible change in the prompt
direction; however, \method achieves this high alignment while being the only approach strictly constrained to the underlying \textsc{FLAME} manifold. ArcFace is
where editing in shape space dominates ($0.93$ vs.\ $0.76$ and
$0.52$): InstructPix2Pix drifts identity, and
GaussianAvatar-Editor's degrades identity preservation; \method
strictly constrains the displacement of $\boldsymbol{\beta}$ within a $300$-D linear subspace
fitted to real heads, ensuring the edited identity is valid by
construction. The lowest $TV_{2D}$ ($0.0122$) reflects that our
edits never pass through a per-frame 2D denoiser, eliminating the
high-frequency residue diffusion-based editors leave on cheeks and
hair silhouette (Fig.~\ref{fig:vs_baselines}).

\begin{table}[htbp]
  \vspace{-0.7cm}
  \centering
  \caption{\textbf{Ablation of the 3DGS enhancements.}}
  \label{tab:ablation}
  \begin{tabular*}{\columnwidth}{@{\extracolsep{\fill}}lccc@{}}
    \toprule
    Variant & CLIP-D $\uparrow$ & ArcFace $\uparrow$ & $TV_{2D}$ $\downarrow$ \\
    \midrule
    \textbf{\method (full)}         & \textbf{0.0490} & 0.9319 & \textbf{0.0122} \\
    \quad w/o (a) Surface Consist.\ & 0.0481 & 0.9278 & 0.0123 \\
    \quad w/o (b) Smooth-Color tanh & 0.0431 & 0.9318 & 0.0125 \\
    \quad w/o (c) Warped Refining   & 0.0470 & \textbf{0.9335} & 0.0131 \\
    \quad w/o (a)+(b)+(c)           & 0.0437 & 0.9322 & 0.0132 \\
    \bottomrule
  \end{tabular*}
  \vspace{-0.8cm}
\end{table}

\paragraph{Ablation insights.}
Table~\ref{tab:ablation} isolates the impact of each 3DGS enhancement. Removing any of (a)~Surface Consistency, (b)~Smooth-Color tanh, or (c)~Warped Refining hurts at least two reported metrics. Removing all the enhancements (last row) lowers CLIP-D and increases $TV_{2D}$, consistent with the stronger floaters seen in Fig.~\ref{fig:ablation}. Each of the 3DGS additions targets a different failure mode: removing \emph{(a) Surface Consistency} costs the most ArcFace as splats detach from the edited mesh; removing \emph{(b) Smooth-Color tanh} is the most damaging on CLIP-D ($-12\%$) because the abrupt sigmoid introduces color transitions that locally disagree with the prompt direction; and removing \emph{(c) Warped Refining} produces the largest $TV_{2D}$ increase and nominally improves ArcFace ($0.9335$) only because, without re-optimization, the avatar barely moves - ArcFace must therefore be read jointly with the other two metrics.

\begin{table}[htbp]
  \vspace{-0.7cm}
    \centering
    \caption{\textbf{Runtime comparison.}}
    \label{tab:timing-avatar}
    \small
    \begin{tabular*}{\columnwidth}{@{\extracolsep{\fill}}lccc@{}}
        \toprule
        Method & Reconstruction$\downarrow$ & Edit$\downarrow$ & Render$\uparrow$ \\
        \midrule
        IPix2Pix~\cite{brooks2022instructpix2pix} & N/A & $\sim$400 ms & N/A \\
        GAEditor~\cite{liu2025gaussianavatar-editor} & \multicolumn{2}{c}{\quad \leftarrowfill \quad $\sim$800 s \quad \rightarrowfill \quad} & $>$100 fps \\
        \textbf{\method (Ours)} & $\sim$60 s & $\sim$30 s & $>$800 fps \\
        \bottomrule
    \end{tabular*}
  \vspace{-0.8cm}
\end{table}

\paragraph{Runtime comparison.}
As detailed in Table~\ref{tab:timing-avatar}, \method's reconstruction is a one-time $\sim$60\,s pass; subsequent edits take $\sim$30\,s each. In contrast, GaussianAvatar-Editor~\cite{liu2025gaussianavatar-editor} couples training and editing into a single $\sim$800\,s optimization per prompt, while InstructPix2Pix~\cite{brooks2022instructpix2pix} processes each frame independently via diffusion and produces no animatable avatar.

% =============================================================================

\subsection{Qualitative results}
\label{sec:experiments:qualitative}

\begin{figure}[htbp]
  \vspace{-0.5cm}
  \centering
  \includegraphics[width=\textwidth]{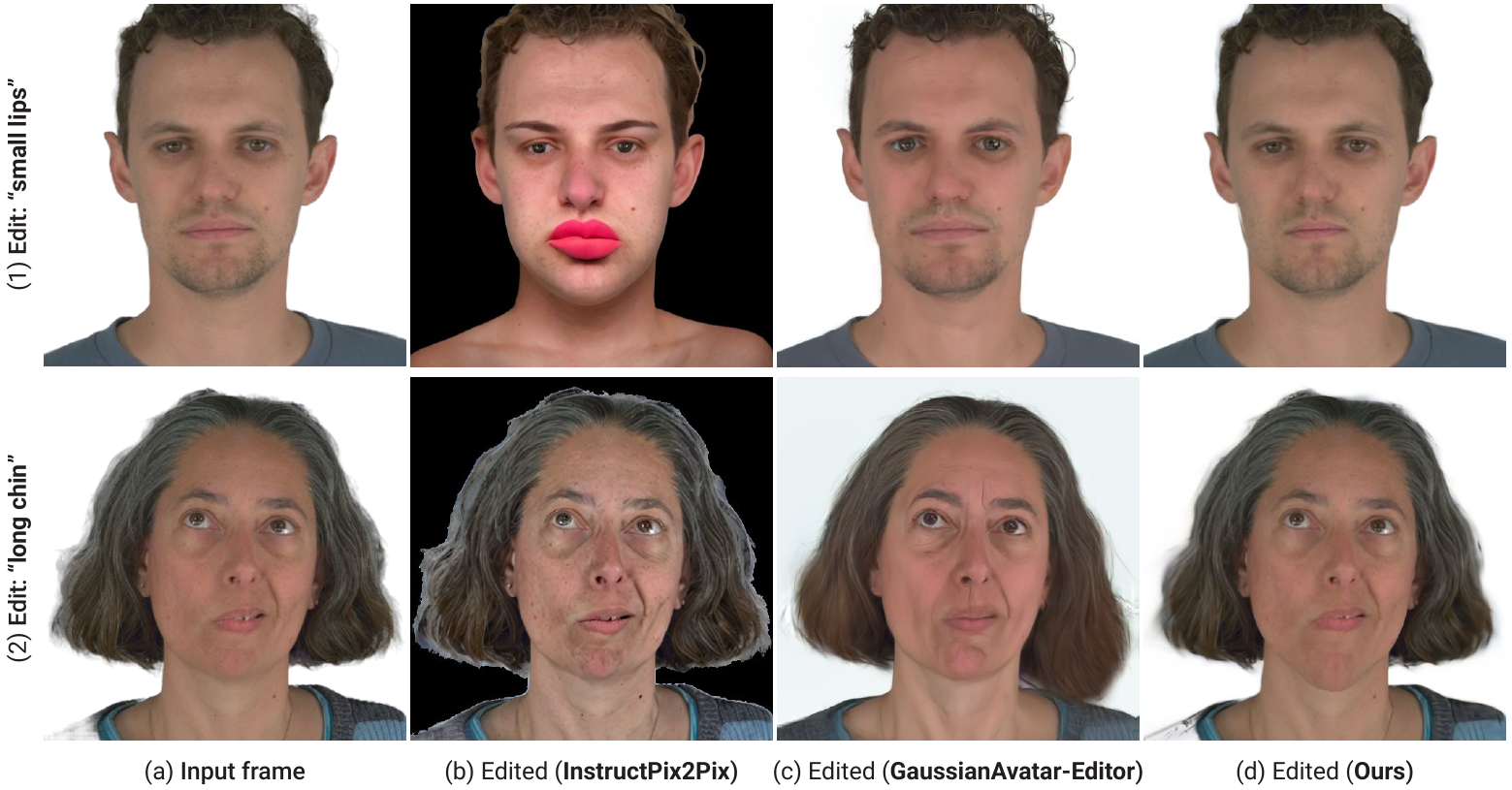}
  \caption{\textbf{Full-avatar editing comparison} on
  \textsc{NeRSemble}~\cite{nersemble} against InstructPix2Pix and
  GaussianAvatar-Editor.}
  \label{fig:vs_baselines}
  \vspace{-0.9cm}
\end{figure}

\paragraph{Full-avatar editing comparison.}
Figure~\ref{fig:vs_baselines} compares the three full-avatar
editors: InstructPix2Pix produces 2D artifacts and identity drift;
GaussianAvatar-Editor preserves structure but barely realizes the
requested geometric change; \method delivers a localized edit while
preserving identity and surrounding appearance.

\begin{figure}[htbp]
  \vspace{-0.5cm}
  \centering
  \includegraphics[width=\textwidth]{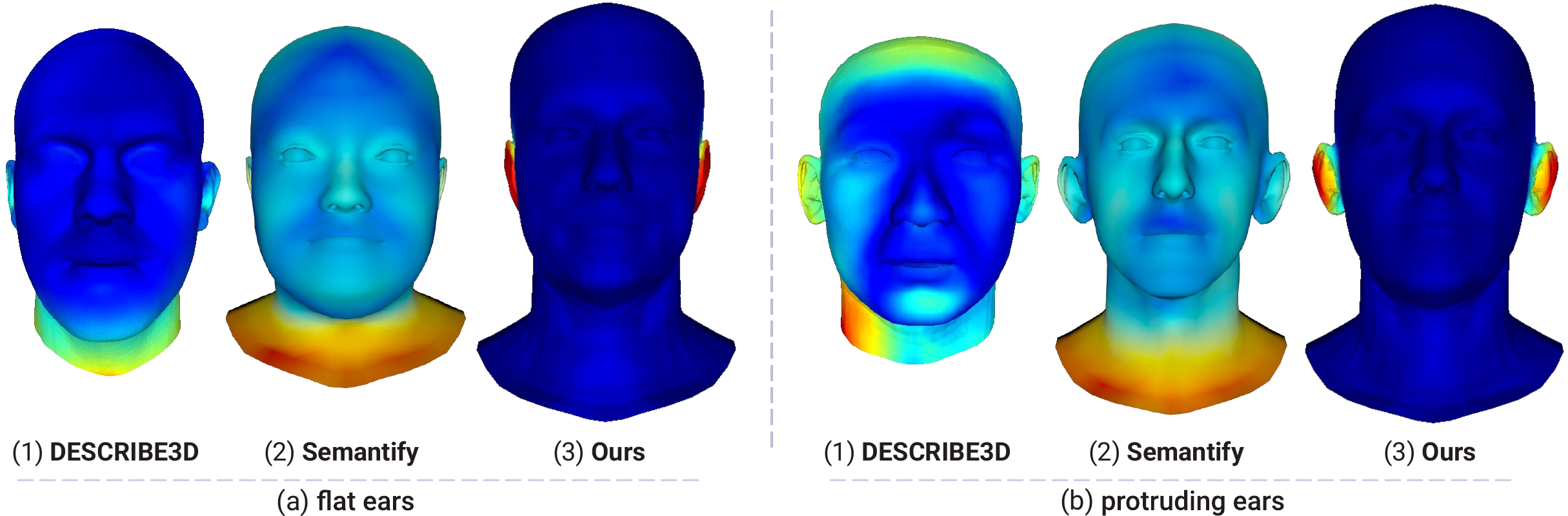}
  \caption{\textbf{Shape editing comparison}. Meshes are color-coded by per-vertex displacement from the
  neutral identity; only \shapemodel~(Ours) keeps the
  displacement concentrated on the ears.}
  \label{fig:shape_vs_baselines}
  \vspace{-0.9cm}
\end{figure}

\paragraph{Shape editing comparison.}
Figure~\ref{fig:shape_vs_baselines} compares
\shapemodel against \mbox{DESCRIBE3D} and the slider-controlled Semantify
on a neutral identity for two ear edits: \mbox{DESCRIBE3D} spreads
displacement over face and neck, Semantify drags the bust and
face along, and only \shapemodel concentrates deformation on the
named region. We report this comparison qualitatively because both
baselines synthesize an identity from scratch rather than editing a
source one, so per-vertex distance metrics such as Chamfer would
conflate edit locality with identity drift and unfairly reward
methods that do not preserve the source identity.

\begin{figure}[htbp]
  \vspace{-0.5cm}
  \centering
  \includegraphics[width=\textwidth]{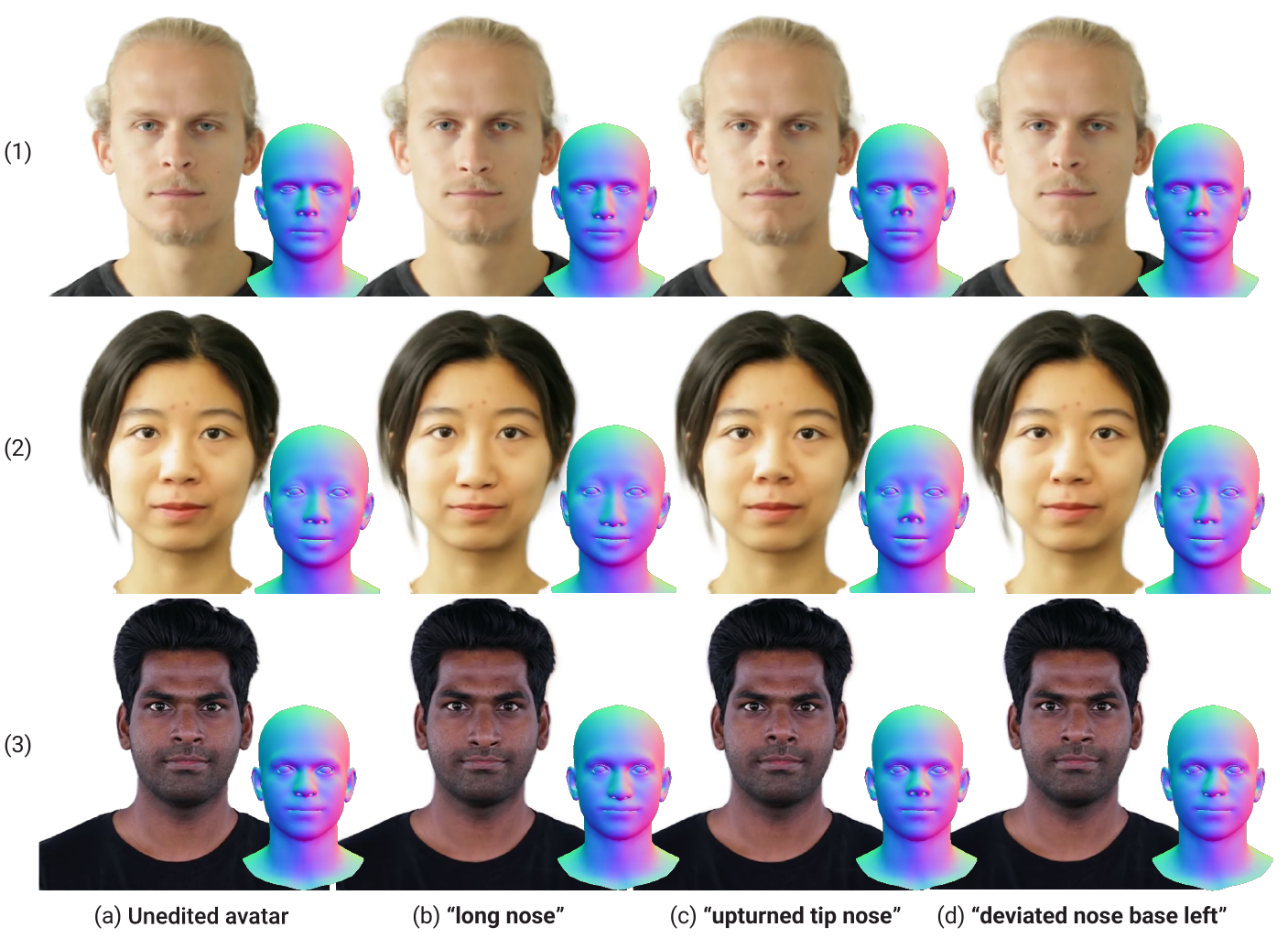}
  \caption{\textbf{Fine-grained local edits.} Illustrates
  \method's ability to realize a wide variety of fine-grained shape
  changes within a single facial region.}
  \label{fig:qualitative}
  \vspace{-0.9cm}
\end{figure}

\paragraph{Fine-grained local edits.}
Figure~\ref{fig:qualitative} shows \method on three subjects across
three nose-region prompts that range from a global elongation
(\emph{``long nose''}) to highly localized tweaks of the tip
(\emph{``upturned tip nose''}) and base (\emph{``deviated nose base
left''}); the normal-coded \textsc{FLAME} inset confirms that the
geometric change is confined to the named region while hair, eyes,
mouth and skin remain untouched. In the supplementary material we
further show that these fine-grained nose edits stay consistent under
non-frontal, near-profile viewpoints on an in-the-wild sequence.

\begin{figure}[htbp]
  \vspace{-0.5cm}
  \centering
  \includegraphics[width=\textwidth]{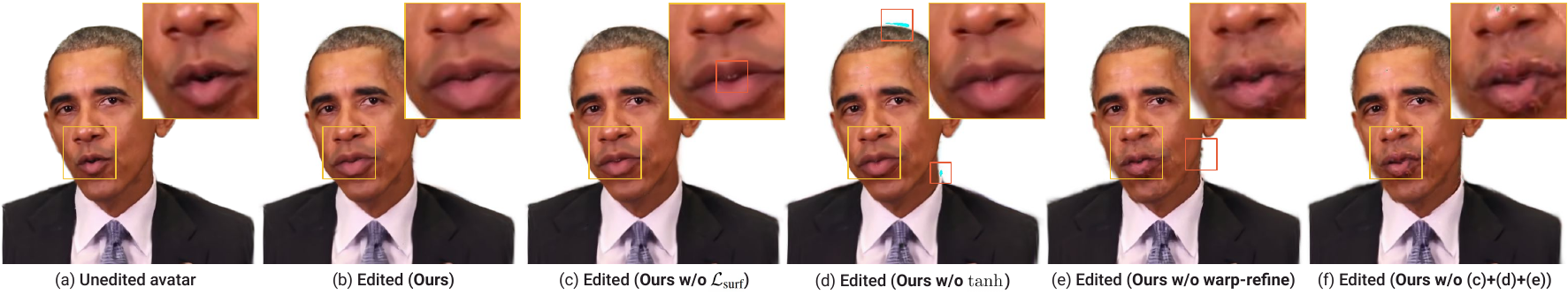}
  \caption{\textbf{Ablation of the 3DGS enhancements} on
  \emph{``thicker lips''}. Red boxes mark the characteristic failure modes
  that each component prevents (floaters and stray colored splats).}
  \label{fig:ablation}
  \vspace{-0.9cm}
\end{figure}

\paragraph{Qualitative ablation.}
Figure~\ref{fig:ablation} visually isolates each 3DGS enhancement on the prompt
\emph{``thicker lips''}: removing $\mathcal{L}_{\text{surf}}$
produces a floater at the lower-lip boundary, reverting to the
sigmoid re-introduces stray colored splats, and skipping warped
refining leaves ghosting and color seams around the lips.

% =============================================================================

\subsection{Limitations}
\label{sec:experiments:limitations}
An inherent challenge is the geometric discrepancy between the parametric \textsc{FLAME} model and the 3D Gaussian representation. While surface-consistency regularization improves this correspondence, perfect topological alignment remains difficult, and residual misalignments can amplify rendering artifacts under extreme shape deformations.

Furthermore, the edit space is bounded by our finite library of predefined deformation prototypes; the model cannot generalize to completely novel structural changes outside this training distribution. Supporting genuinely unseen deformations still requires manual annotation, making automated, data-driven prototype extraction an interesting future direction. Finally, our edits are strictly limited to the \textsc{FLAME} surface geometry; external regions like hair, alongside appearance attributes (e.g., skin or eye color), are not addressed in this work.

\section{Conclusions}
\label{sec:conclusions}

We introduced \method, the first end-to-end pipeline
for language-guided semantic shape editing of animatable 3D Gaussian
head avatars reconstructed from monocular video. By decoupling text-driven geometric manipulation via 
\shapemodel from the underlying splat-based rendering, we achieve localized, 
identity-preserving edits without compromising animation fidelity. 
Furthermore, our targeted rendering regularizations ensure that the 
avatar remains visually coherent under structural deformations,
reducing each edit to a brief refining of the pre-trained avatar
instead of a costly full re-optimization. Ultimately, 
\method unfreezes the geometric identity of monocular avatars and lays 
the groundwork for fully conversational, real-time avatar sculpting. 
This bridges a critical gap in virtual telepresence and dynamic media, 
bringing us a step closer to highly responsive, language-driven 3D 
digital humans.

\section{Acknowledgments}
\label{sec:acknowledgments}

This work has been partially supported by project {\small PID2023-151184OB-I00}, funded by the Ministerio de Ciencia e Innovación (MCIN) / Agencia Estatal de Investigación (AEI) / 10.13039/501100011033, by the European Social Fund (ESF) ``Investing in your future'' and by \textit{ERDF, EU}; and by the industrial doctorate 2021 DI 00102 funded by the Government of Catalonia.

% -----------------------------------------------------------------------------
% Supplementary material
% -----------------------------------------------------------------------------
\clearpage
\appendix

\section{Implementation Details}
\label{sec:supp:impl}

We detail the architecture of the \shapemodel network, the generation
process for our synthetic paired dataset, and the hyper-parameters
governing the 3D Gaussian Splatting optimization.

% =============================================================================
\subsection{\shapemodel architecture}
\label{sec:supp:arch}

% Fig.~\ref{M-fig:method}
% \textbf{Fig.~2}

\paragraph{Text and shape encoders.} As illustrated in the bottom
panel of \textbf{Fig.~2} in the main manuscript, the prompt is encoded by a
frozen Flan-T5-Large~\cite{chung2024flan,raffel2020t5} encoder,
yielding $1024$-D token features
$\mathbf{T}\!\in\!\mathbb{R}^{L\times1024}$ (we cap $L{=}128$ tokens)
and a masked-mean sentence embedding
$\mathbf{t}_g\!\in\!\mathbb{R}^{1024}$. The source shape
$\boldsymbol{\beta}_0\!\in\!\mathbb{R}^{300}$ is projected by a small
MLP (hidden sizes $[512,512]$, LayerNorm, dropout $0.15$) to a single
$512$-D token $\mathbf{s}\!\in\!\mathbb{R}^{512}$.

\paragraph{Cross-attention fusion.} A multi-head attention block
($d{=}512$, $8$ heads, attention dropout $0.1$) treats $\mathbf{s}$ as
the query and the linearly projected text tokens
($\mathbb{R}^{1024}\!\to\!\mathbb{R}^{512}$) as keys/values,
preserving the T5 padding mask. A residual connection and LayerNorm
yield a fused embedding $\mathbf{h}\!\in\!\mathbb{R}^{512}$ carrying
shape-conditioned semantic context.

\paragraph{Gate network.} Conditioning the gate solely on the global
text embedding $\mathbf{t}_g$ forces the network to localize the edit
strictly from language. A 1-hidden-layer MLP (hidden size $512$,
dropout $0.15$) maps a separate
$\mathbb{R}^{1024}\!\to\!\mathbb{R}^{512}$ projection of $\mathbf{t}_g$
to $K{=}6$ logits and applies a softmax, producing
$\mathbf{g}\!\in\!\Delta^{K-1}$ over the \textsc{FLAME} regions
\{\emph{face, nose, lips, ears, neck, eyes}\}.

\paragraph{Prior and posterior.} Two MLP heads (hidden sizes
$[768,512]$, LayerNorm, dropout $0.15$) parameterize diagonal
Gaussian distributions over a structured latent
$\mathbf{z}\!\in\!\mathbb{R}^{K d}$ with $K{=}6$ and $d{=}8$
(so $\dim\mathbf{z}{=}48$): the prior
$p_\theta(\mathbf{z}\!\mid\!\mathbf{t},\boldsymbol{\beta}_0)
=\mathcal{N}(\boldsymbol{\mu}_p,\boldsymbol{\sigma}_p^2)$ is regressed
from $\mathbf{h}$, while the training-only posterior
$q_\phi(\mathbf{z}\!\mid\!\mathbf{t},\boldsymbol{\beta}_0,
\Delta\boldsymbol{\beta})=\mathcal{N}(\boldsymbol{\mu}_q,
\boldsymbol{\sigma}_q^2)$ takes
$[\mathbf{h};\,E_\Delta(\Delta\boldsymbol{\beta})]$ as input, where
$E_\Delta$ is a $[512,512]$ MLP encoder of the ground-truth shape
delta. $\mathbf{z}$ is sliced into $K$ contiguous $d$-dimensional
blocks, one per attribute.

\paragraph{Decoder.} The latent is gated per region,
$\tilde{\mathbf{z}}_k = g_k\,\mathbf{z}_k$, flattened, and
concatenated with the source shape embedding $\mathbf{s}$ (passed
through dropout $p{=}0.3$ to discourage the decoder from shortcutting
through source identity and bypassing $\mathbf{z}$). A 4-layer MLP
$D$ (hidden sizes $[768,512,512]$, dropout $0.15$) then produces the
shape delta:
\begin{equation}
  \Delta\boldsymbol{\beta} \;=\;
  D\Big( \big[ g_k\,\mathbf{z}_k \big]_{k=1}^K \;\big\Vert\; \mathbf{s} \Big) \in \mathbb{R}^{300}.
  \label{eq:supp:decoder}
\end{equation}
The predicted edited shape is then naturally obtained as
$\boldsymbol{\beta}_1^{\text{pred}} = \boldsymbol{\beta}_0 +
\Delta\boldsymbol{\beta}$, which is subsequently used to compute the
geometric training objectives.

\paragraph{Inference.} At test time we drop the posterior and use the
prior mean rescaled by a prior-scale factor,
$\mathbf{z}=s\,\boldsymbol{\mu}_p$ with $s{=}1.2$, which compensates
for the natural attenuation of $\boldsymbol{\mu}_p$ relative to a
sample from $q_\phi$ during training. The mapping is deterministic
per prompt; varying $s$ continuously modulates edit strength.

\paragraph{Optimization.} \shapemodel is trained with AdamW
(learning rate $10^{-4}$, weight decay $0.01$, gradient-norm
clipping at $1.0$) for $100$ epochs at batch size $256$ on a single
NVIDIA RTX~5090.

% =============================================================================
\subsection{Training data}
\label{sec:supp:data}

We synthesize triplets
$(\boldsymbol{\beta}_0,\mathbf{t},\boldsymbol{\beta}_1)$ from $300$
MICA-fitted LYHM~\cite{lyhm} identities. For each of the six
\textsc{FLAME} regions we author $\sim$130 fine-grained edit
templates specified as small sets of canonical vertex indices paired
with 3D displacement vectors (\emph{nose}: short/long, narrow/wide,
deviated, bridge/tip variants, etc.; analogous for the other
regions). At each scale $s\!\in\!\{0.5,1.0,1.5\}$ and for every
(subject, edit template) pair we recover the corresponding $\Delta\boldsymbol{\beta}^{*}$ by an
iterative projection (Adam, $1500$ steps, lr $10^{-2}$,
$\ell_2$ regularization weight $10^{-6}$) onto the \textsc{FLAME}
shape basis with an outside-region penalty
($\lambda_{\text{out}}{=}10^{2}$) that keeps the solution from
leaking into unrelated areas. Each
$(\boldsymbol{\beta}_0,\text{edit},\text{scale})$ tuple is paired
with five template-paraphrased prompts whose qualifiers
(``\emph{very subtly}'', ``\emph{slightly}'', \ldots,
``\emph{drastically}'') reflect the scale, yielding ${\sim}500$k
supervised triplets. Data generation takes $\sim$70~hours on a single RTX~5090, and \shapemodel training takes $\sim$40~hours on the same GPU.
$15\%$ of the triplets is held out for validation.

% =============================================================================
\subsection{Full training objective}
\label{sec:supp:loss}

The full objective combines vertex-space supervision, parameter
supervision, an ELBO term, and the gate cross-entropy:
\begin{align}
  \mathcal{L} \;=\;
    & \lambda_v^{\text{in}}\,\|V(\boldsymbol{\beta}_1^{\text{pred}})\!-\!V(\boldsymbol{\beta}_1)\|^2_{V_{a^{*}}}
    \notag\\
    & +\;\lambda_v^{\text{out}}\,\|V(\boldsymbol{\beta}_1^{\text{pred}})\!-\!V(\boldsymbol{\beta}_0)\|^2_{V\setminus V_{a^{*}}}
    \notag\\
    & +\;\lambda_{\boldsymbol{\beta}}\,\|\boldsymbol{\beta}_1^{\text{pred}}\!-\!\boldsymbol{\beta}_1\|^2
    \;+\;\lambda_g\,\textsc{CE}(\mathbf{g},a^{*})\notag\\
    & +\;\lambda_{kl}\,w(t)\sum_{i=1}^{Kd}
        \max\!\big(\textsc{FB},\,
        \textsc{KL}\!\big(q_\phi^{(i)}\,\|\,p_\theta^{(i)}\big)\big)\,,
  \label{eq:supp:loss}
\end{align}
where $a^{*}$ is the ground-truth region, $V_{a^{*}}$ are its
vertices, $V(\cdot)$ denotes the \textsc{FLAME} forward at zero
expression/pose, $w(t)$ is a linear KL warm-up schedule, and
$\textsc{FB}$ is a per-dimension free-bits floor that prevents
posterior collapse without distorting the prior. The two vertex
terms decompose the supervision into the region where the edit
should occur (toward the target mesh) and its complement,
anchored to the source. We use
$\lambda_v^{\text{in}}{=}\lambda_v^{\text{out}}{=}5\!\times\!10^{5}$,
$\lambda_{\boldsymbol{\beta}}{=}10$,
$\lambda_g{=}10$,
$\lambda_{kl}{=}0.05$,
$\textsc{FB}{=}0.03$, with $w(t)$ ramping linearly from $0$ to $1$
over the first $10$ epochs.

% =============================================================================
\subsection{Warped refining: per-pixel resampling map}
\label{sec:supp:warp}

Let $M_0(\boldsymbol{\beta}_0)$ and $M_1(\boldsymbol{\beta}_1)$ be
the unedited and edited meshes; for each training camera and frame
$t$ we (i) render the base avatar with $M_0$ to obtain
$\tilde I_t$ and (ii) rasterize $M_1$ to get, at every pixel $p$
that the edited mesh covers, the originating triangle $f(p)$ and its
barycentric coordinates $\boldsymbol{\lambda}(p)\!\in\!\Delta^2$.
The exact source pixel to sample from is then
$\pi\big(\sum_{k}\lambda_k(p)\,V^{(0)}_{f(p),k}\big)$, the
reprojection through the same camera $\pi$ of the point with the
same triangle/barycentric on the unedited mesh $V^{(0)}$.
This is a per-pixel resampling map $W_t$, and the view-consistent
pseudo-ground-truth $\hat I_t = W_t(\tilde I_t)$ is obtained by a
single \texttt{remap} (pixels outside the edited mesh fall back to
identity sampling). Because the warp is derived from the shared
mesh topology rather than from image-space optical flow, it is
insensitive to texture, exact under the rasterization of $M_1$, and
free of the bracketing failures that optical-flow estimators
exhibit in low-texture regions such as cheeks or forehead.

The anchor term in the refining stage,
\begin{equation}
  \mathcal{L}_{\text{anchor}} \;=\;
  \frac{1}{|\Phi|}\sum_{\phi\in\Phi}\|\phi-\phi^{(0)}\|_2^2,
  \label{eq:supp:anchor}
\end{equation}
prevents the non-frozen parameters $\Phi$ (positions, opacities,
scales, rotations and the position/rotation blendshape bases) from
drifting away from the warm start $\Phi^{(0)}$; we use
$\lambda_{\text{anchor}}{=}10^{-2}$ and freeze the spherical-harmonic
DC color channels throughout refining.

% =============================================================================
\subsection{3DGS backbone and refining hyper-parameters}
\label{sec:supp:gs}

% (Eq.~\ref{M-eq:surf})
% (\textbf{Eq.~2})

Both the base reconstruction and the refine pass share the
same \mbox{RGBAvatar~\cite{li2025rgbavatar}} backbone with
$B{=}20$ identity blendshapes layered on top of the $129$ FLAME bases,
$256$-resolution texture, isotropic scale initialization
$0.8\!\times\!10^{-3}$ and opacity initialization $0.5$. We train
for $50{,}000$ iterations at batch size $10$ with the Adam learning
rates of the original \mbox{RGBAvatar} schedule
($\text{position}{=}8\!\times\!10^{-4}$, exponentially decayed to
$8\!\times\!10^{-6}$ over $30{,}000$ steps;
$\text{opacity}{=}0.25$;
$\text{scale}{=}2.5\!\times\!10^{-2}$;
$\text{rotation}{=}5\!\times\!10^{-3}$;
SH features $1.25\!\times\!10^{-2}$).
Blendshape branches are activated after $3{,}000$ warm-up iterations.
The surface-consistency regularizer of the main paper
(\textbf{Eq.~2}) corresponds to $\lambda_{x}{=}3$ and
$\lambda_{\Delta x}{=}1$ in our notation.
Warped refining is initialized from the base-reconstruction
checkpoint of the same subject (which is trained once per identity and shared
across all of its edits) and reuses the same configuration with the
SH-DC color channels frozen, the anchor loss enabled, and the
training images replaced by the warped pseudo-ground-truths
$\{\hat I_t\}$.

% =============================================================================
\subsection{Evaluation metrics}
\label{sec:supp:metrics}

We report three complementary metrics that jointly probe prompt
alignment, identity preservation, and rendering quality. Let
$I^{\text{src}}_n$ and $I^{\text{edit}}_n$ denote the source frame
and our edited render for frame $n$, $\mathbf{t}^{\text{src}}$ and
$\mathbf{t}^{\text{edit}}$ the source and edit prompts, and
$N$ the number of evaluation frames.

\paragraph{CLIP directional similarity (CLIP-D)~\cite{clip,gal2021stylegannada}.}
Building upon the text-driven semantic manipulation established by
StyleCLIP~\cite{patashnik2021styleclip}, we evaluate semantic
alignment using the Directional CLIP Similarity metric introduced by
Gal~\etal in StyleGAN-NADA~\cite{gal2021stylegannada}. Rather than
the absolute image-text similarity (which is dominated by
identity-agnostic content), CLIP-D measures whether the
visual change between source and edit aligns with the
textual change between source and edit prompts in the CLIP
ViT-B/16 feature space:
\begin{equation}
  \Delta\mathbf{T} = \mathcal{E}_T(\mathbf{t}^{\text{edit}}) - \mathcal{E}_T(\mathbf{t}^{\text{src}}),\quad
  \Delta\mathbf{I}_n = \mathcal{E}_I(I^{\text{edit}}_n) - \mathcal{E}_I(I^{\text{src}}_n),
  \label{eq:supp:clipd_deltas}
\end{equation}
\begin{equation}
  \textsc{CLIP-D} \;=\;
    \left|\;\frac{1}{N}\sum_{n=1}^{N}
      \frac{\langle\Delta\mathbf{I}_n,\Delta\mathbf{T}\rangle}
           {\|\Delta\mathbf{I}_n\|\,\|\Delta\mathbf{T}\|}
    \;\right|,
  \label{eq:supp:clipd}
\end{equation}
with $\mathcal{E}_I,\mathcal{E}_T$ the CLIP image and text encoders.
Reporting $\bigl|\overline{\text{dirs}}\bigr|$ keeps the aggregated
score positive and prevents sign cancellation across frames whose
camera or expression briefly flips the projection's sign while the
average direction remains aligned with the prompt. Higher is better.

\paragraph{ArcFace identity similarity~\cite{deng2019arcface}.}
For each frame, we detect a centered face and use the same five-point
landmarks to align both $I^{\text{src}}_n$ and $I^{\text{edit}}_n$ to
a canonical $112\!\times\!112$ crop, extract $\ell_2$-normalized
$512$-D ArcFace embeddings $\mathbf{f}^{\text{src}}_n$ and
$\mathbf{f}^{\text{edit}}_n$, and report the average cosine
similarity:
\begin{equation}
  \textsc{ArcFace} \;=\;
    \frac{1}{N}\sum_{n=1}^{N}
      \frac{\langle\mathbf{f}^{\text{src}}_n,\mathbf{f}^{\text{edit}}_n\rangle}
           {\|\mathbf{f}^{\text{src}}_n\|\,\|\mathbf{f}^{\text{edit}}_n\|}.
  \label{eq:supp:arcface}
\end{equation}
This isolates identity drift from texture and pose differences and is
the strict-sense identity-preservation metric we report. Higher is
better.

\paragraph{Anisotropic 2D Total Variation ($TV_{2D}$).}
We use $TV_{2D}$ as a no-reference image-quality proxy that
penalizes high-frequency spatial artifacts - floaters, jagged color
seams, stair-stepped edges along the silhouette - characteristic of
splat-based renderings with misaligned Gaussians. For an RGB frame
$I\!\in\![0,1]^{H\times W\times3}$,
\begin{equation}
  \mathrm{TV}_{2D}(I) \;=\;
    \frac{1}{|\Omega_y|}\!\sum_{(i,j,c)\in\Omega_y}\!|I_{i+1,j,c}\!-\!I_{i,j,c}|
  \;+\;
    \frac{1}{|\Omega_x|}\!\sum_{(i,j,c)\in\Omega_x}\!|I_{i,j+1,c}\!-\!I_{i,j,c}|,
  \label{eq:supp:tv2d}
\end{equation}
where $\Omega_y$ and $\Omega_x$ are the sets of valid vertical and
horizontal pixel pairs across all channels, and the average is
finally taken over the $N$ evaluation frames. Normalizing by pixel
count makes the score scale-comparable across resolutions; lower is
better.

\section{Additional Results}
\label{sec:supp:addresults}

% =============================================================================
\subsection{Per-region quantitative breakdown}
\label{sec:supp:perregion}

\begin{table}[htbp]
  \vspace{-0.5cm}
  \centering
  \caption{\textbf{Per-region quantitative results of \method.}}
  \label{tab:supp:perregion}
  \begin{tabular*}{\columnwidth}{@{\extracolsep{\fill}}lccc@{}}
    \toprule
    Region & CLIP-D $\uparrow$ & ArcFace $\uparrow$ & $TV_{2D}$ $\downarrow$ \\
    \midrule
    Face  & 0.0440 & 0.8879 & 0.0121 \\
    Nose  & 0.0498 & 0.9329 & 0.0122 \\
    Eyes  & 0.0525 & 0.9060 & 0.0122 \\
    Ears  & 0.0745 & 0.9741 & 0.0123 \\
    Neck  & 0.0448 & 0.9642 & 0.0123 \\
    Lips  & 0.0282 & 0.9262 & 0.0123 \\
    \midrule
    Mean & 0.0490 & 0.9319 & 0.0122 \\
    \bottomrule
  \end{tabular*}
  \vspace{-0.3cm}
\end{table}

% Table~\ref{M-tab:avatar}
% \textbf{Table~1}

Table~\ref{tab:supp:perregion} disaggregates the aggregate scores of
\textbf{Table~1} in the main paper across the six \textsc{FLAME} regions edited
by \method. The $TV_{2D}$ column is essentially flat
($\sigma{=}8\!\times\!10^{-5}$), confirming that the warped-refining
stage and the surface-consistency regularizer keep rendering quality
region-agnostic regardless of the edited attribute.
Identity preservation correlates with the spatial overlap of each
region with the front-facing portrait used by ArcFace: peripheral
regions (\emph{ears}, \emph{neck}) leave the identity-bearing inner
face untouched and score highest, while direct reshaping of core
features (\emph{face}, \emph{eyes}) naturally induces a larger
identity shift - yet even in the most severe case ArcFace remains
above $0.88$, well clear of the $0.76$ obtained by
GaussianAvatar-Editor on full edits. \mbox{CLIP-D}, in turn, reflects the
silhouette footprint of the edit rather than its semantic fidelity:
peripheral regions admit larger contour changes than highly localized
prompts on \emph{lips}, so they project further onto the textual edit
direction. This is a known property of CLIP-D~\cite{gal2021stylegannada}
and is consistent with the qualitative locality shown in the heatmaps
of Fig.~\ref{fig:supp:heatmaps}.

% =============================================================================

\subsection{Edit-locality heatmaps}
\label{sec:supp:heatmaps}

\begin{figure}[htbp]
  \vspace{-0.2cm}
  \centering
  \includegraphics[width=\textwidth]{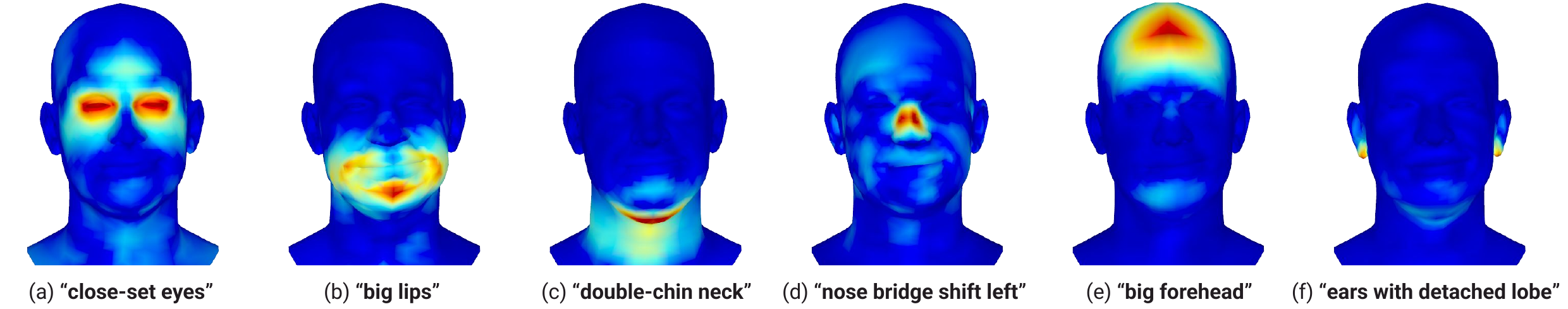}
  \caption{\textbf{Per-vertex displacement heatmaps for representative
  \shapemodel edits.} Color encodes the $\ell_2$ norm of each
  vertex's displacement between the edited and unedited \textsc{FLAME}
  meshes (blue $\to$ red), normalized per mesh. Each subfigure is
  labeled with the prompt fragment fed to \shapemodel. All heatmaps are computed using the source identity of subject \texttt{104\_EXP\_5} from the NeRSemble dataset~\cite{nersemble}.}
  \label{fig:supp:heatmaps}
  \vspace{-0.3cm}
\end{figure}

% (Sec.~\ref{M-sec:method:shapemodel})
% (\textbf{Sec.~3.3})

Figure~\ref{fig:supp:heatmaps} visualizes the geometric extent of
six representative prompts - one per \textsc{FLAME} region - as
per-vertex displacement heatmaps between the edited and unedited
meshes. In every case the energy of the edit concentrates almost
exclusively on the named region: \emph{``close-set eyes''} fires on
the eye-socket and adjacent brow ridge, \emph{``big lips''} on the
upper and lower vermilion border, \emph{``double-chin neck''} on the
submental crease, \emph{``nose bridge shift left''} on the nasal
ridge, \emph{``big forehead''} on the upper frontal bone, and
\emph{``ears with detached lobe''} sharply on the earlobe contour.
The gate-induced disentanglement described in the main paper
(\textbf{Sec.~3.3}) thus translates directly into spatial sparsity in the
\textsc{FLAME} delta itself, leaving adjacent structures undisturbed
and enabling sequential, composable edits without interference. The
locality is moreover semantic rather than rigidly bounded by the
authored vertex masks: the \emph{``nose bridge shift left''} edit
tapers smoothly along the nasal ridge instead of clipping to the
canonical nose mask, indicating that the prior $p_\theta$ has
internalized the correlation structure of the \textsc{FLAME} manifold
rather than mechanically reproducing the authored edit template.

% =============================================================================
\subsection{Extended qualitative results}
\label{sec:supp:results}

\begin{figure}[p]
  \centering
  \includegraphics[width=\textwidth,height=0.85\textheight,keepaspectratio]{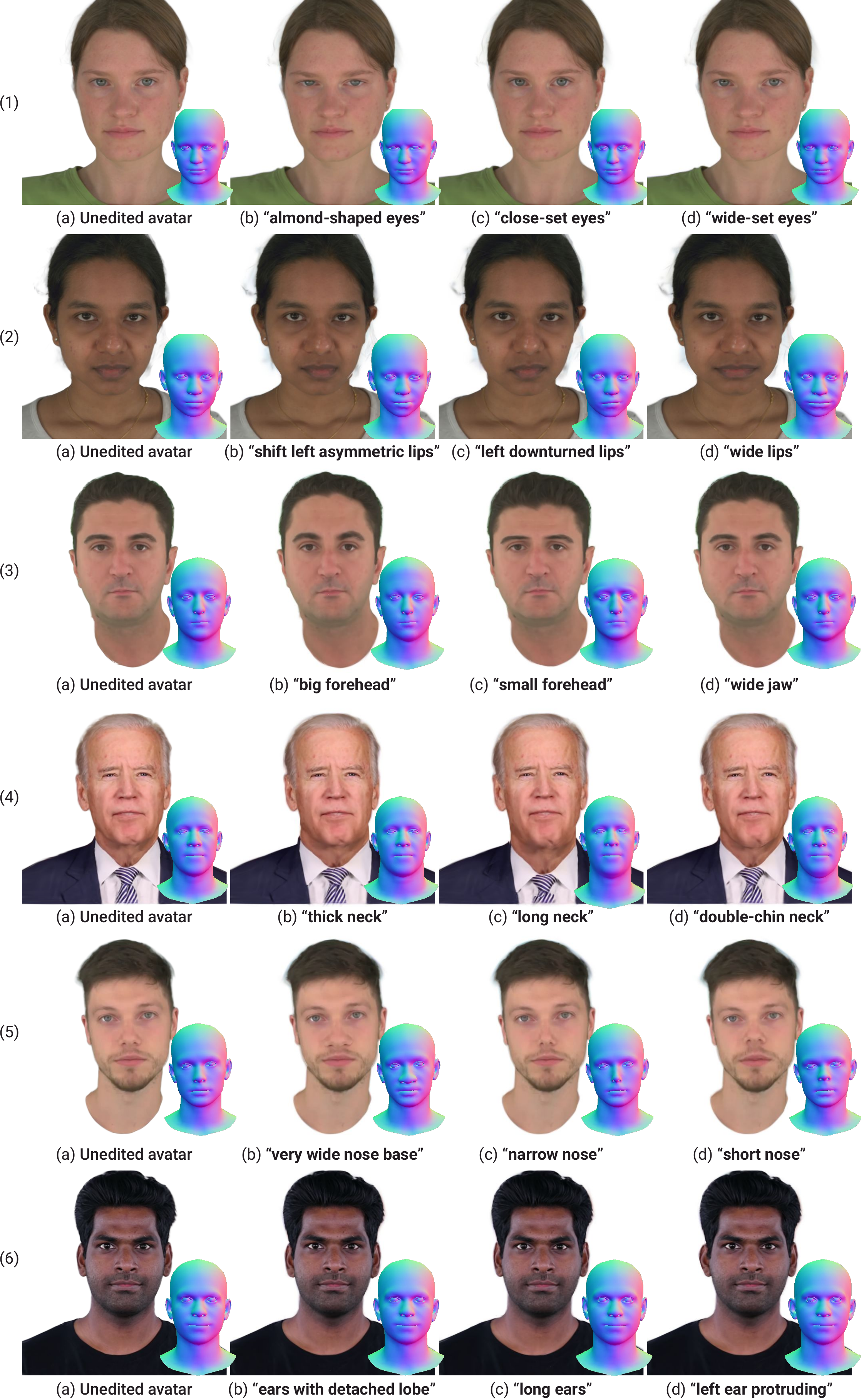}
  \caption{\textbf{Extended qualitative results.} This figure illustrates \method's ability to realize a wide variety of fine-grained shape edits across diverse subjects and attributes.}
  \label{fig:supp:results}
  \vspace{-0.3cm}
\end{figure}

% (Fig.~\ref{M-fig:qualitative})
% (\textbf{Fig.~5})

Figure~\ref{fig:supp:results} extends the qualitative results of the
main paper (\textbf{Fig.~5}) to six subjects covering a broad range of
ethnicities, skin tones, ages, and hairstyles, with three prompts
per subject sampled across the eyes, lips, face, neck, nose and ears
regions. Because the underlying \textsc{FLAME} delta is itself local
(Fig.~\ref{fig:supp:heatmaps}), the warped-refining stage transports
the avatar's appearance along the perturbed geometry instead of
re-synthesizing unedited regions: ``\emph{close-set eyes}'' alters the
inter-ocular distance without touching the lips or hairline, and
``\emph{wide lips}'' stretches the mouth without modifying the nasal
base. The effect persists under intra-region variation - three
distinct prompts on the same subject (e.g.\ row~3: \emph{big
forehead}, \emph{small forehead}, \emph{wide jaw}) yield clearly
different face shapes while preserving eyebrow shape, skin texture
and eye color, which is what drives the high ArcFace scores in
Table~\ref{tab:supp:perregion}. Consistent with our per-region analysis
(Sec.~\ref{sec:supp:perregion}), edits to the ears naturally appear as subtle silhouette
changes from a frontal view, also confirming that the central identity
features remain completely disentangled. Since every edit acts on the
mesh-bound 3D Gaussian avatar rather than on a single 2D image, the
rendered portraits keep their view-dependent appearance and can be
driven by the tracker's expression/pose stream at $>$$800$\,fps; the
normal-coded \textsc{FLAME} insets visualize the underlying geometry
that any downstream animation will inherit.

\begin{figure}[htbp]
  \vspace{-0.2cm}
  \centering
  \includegraphics[width=\textwidth]{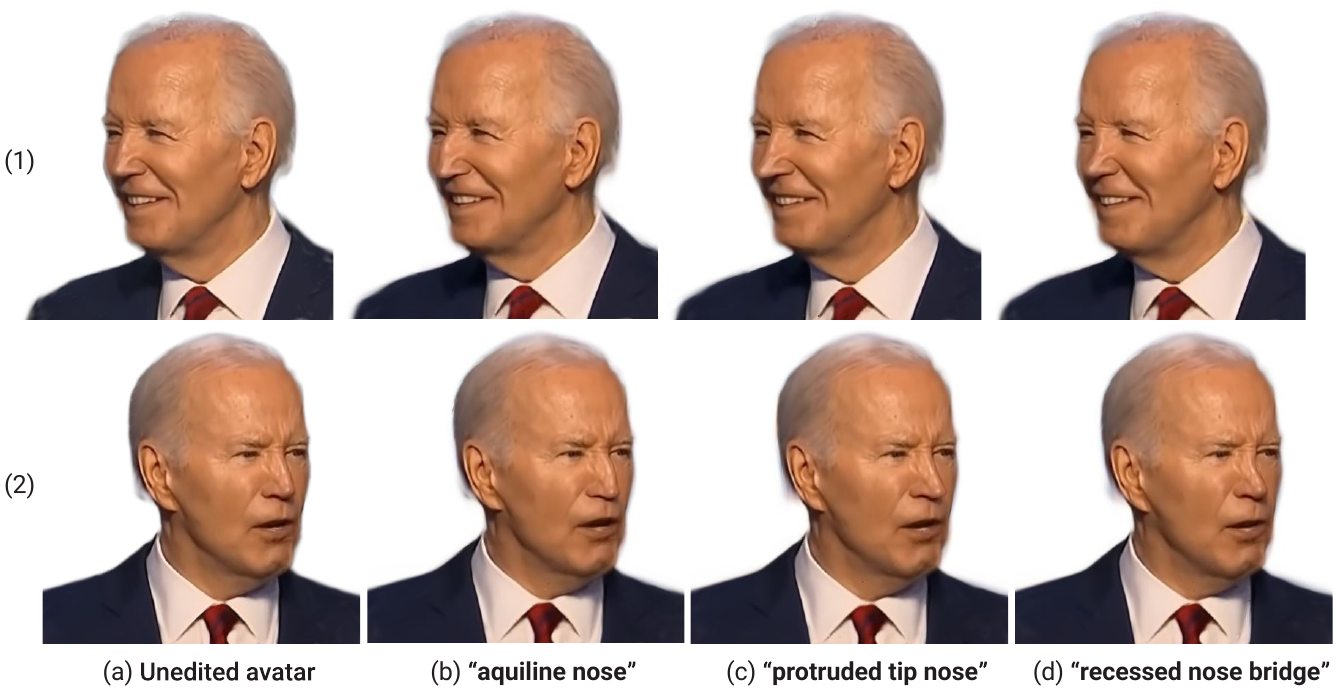}
  \caption{\textbf{Fine-grained nose edits under non-frontal views on an
  in-the-wild sequence.} (a)~the unedited reconstructed avatar and (b-d)~three fine-grained nose edits, all
  rendered from (1-2)~two near-profile viewpoints that best expose the nasal
  geometry.}
  \label{fig:supp:otherviews}
  \vspace{-0.3cm}
\end{figure}

To demonstrate robustness beyond frontal views and controlled environments, Fig.~\ref{fig:supp:otherviews} evaluates \method on an in-the-wild monocular video lacking a studio setup or controlled lighting. Despite this domain shift from the \textsc{NeRSemble}~\cite{nersemble} dataset, the avatar maintains a coherent silhouette and view-dependent shading in near-profile poses. These non-frontal viewpoints are crucial to clearly visualize shape changes in nasal attributes (\emph{``aquiline nose''}, \emph{``protruded tip nose''}, and \emph{``recessed nose bridge''}) that are largely unnoticeable in purely frontal views. In all cases, the deformations are sharply confined to the targeted region, preserving the surrounding identity and face structure, thereby confirming our method's edit locality and multi-view consistency on unconstrained data.

% -----------------------------------------------------------------------------
% Bibliography
% -----------------------------------------------------------------------------
\FloatBarrier

\bibliographystyle{splncs04}
\bibliography{egbib}

\end{document}